\documentclass{vgtc}                          

\PassOptionsToPackage{svgnames}{xcolor}
\usepackage[svgnames,dvipsnames]{xcolor}

\graphicspath{{figures/}{pictures/}{images/}{./}} 

\usepackage{times}                     

\usepackage{tabu}                      
\usepackage{booktabs}                  
\usepackage{lipsum}                    
\usepackage{mwe}                       

\usepackage{mathptmx}                  
\usepackage{amsmath}
\usepackage{amsfonts}
\usepackage{adjustbox}

\usepackage{xcolor}

\onlineid{1181}

\vgtccategory{Technique or Algorithm}

\vgtcinsertpkg

\title{Attention-Guided Saliency Maps for Interpreting Visualization Literacy in VLMs}

\author{
\makebox[0.9\textwidth][c]{%
\begin{minipage}{1.7in}\centering
Maeve Hutchinson\thanks{Both authors contributed equally to this work.}  \footnotemark[2]\\
\scriptsize City St George's,\\University of London
\end{minipage}\hfill
\begin{minipage}{2.1in}\centering
Abderrahmane Wassim Mehdaoui\footnotemark[1]\\
\scriptsize City St George's,\\University of London
\end{minipage}\hfill
\begin{minipage}{1.9in}\centering
Pranava Madhyastha\thanks{Correspondence:~\mbox{\{maeve.hutchinson, pranava.madhyastha\}@city.ac.uk}}\\
\scriptsize City St George's, University of London;\\The Alan Turing Institute
\end{minipage}
}
}

\teaser{
  \begin{adjustbox}{max width=0.87\linewidth, center}
  \includegraphics[width=\linewidth]{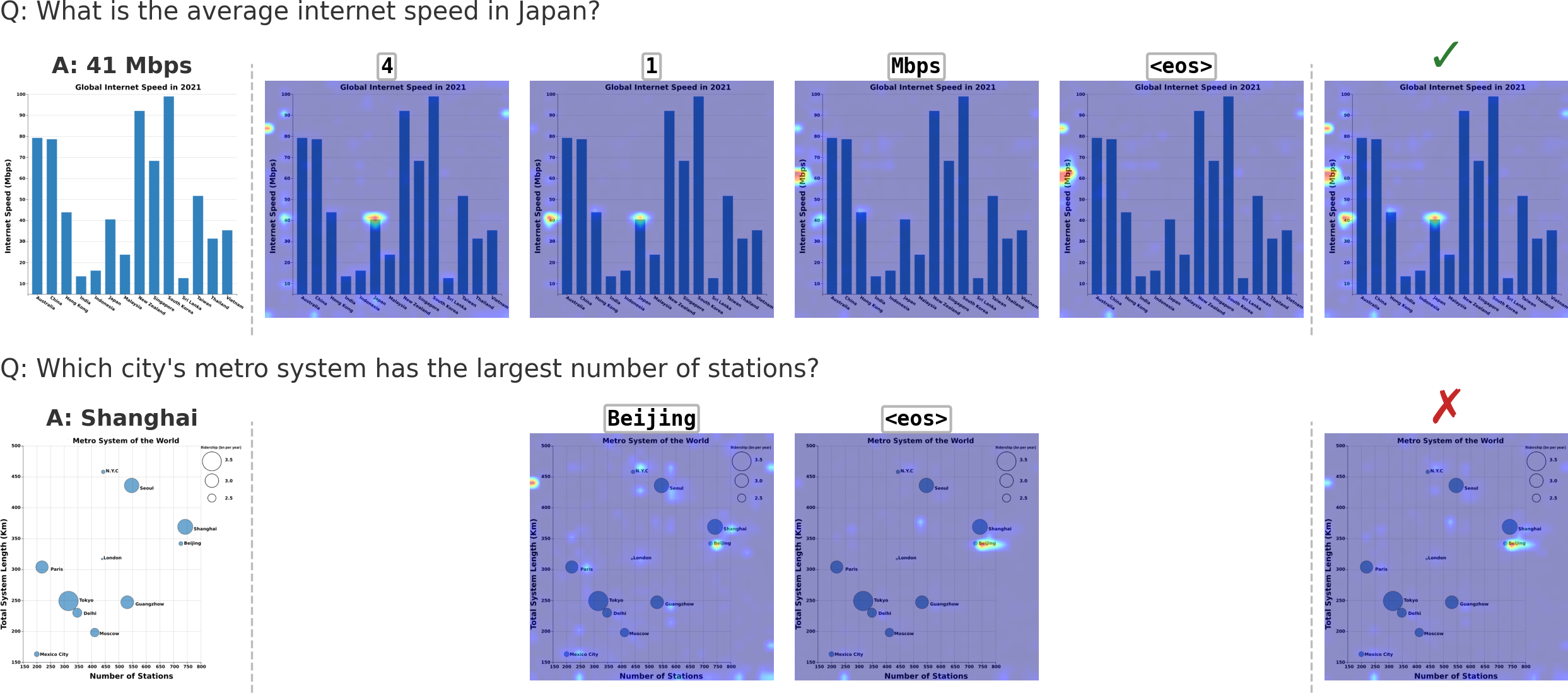}
   \end{adjustbox}
  \caption{Our token-to-patch attention-guided saliency maps for two Mini-VLAT questions answered by ChartGemma. Each row shows, from left to right: the original chart image with the ground-truth answer (A), the saliency map for each generated token, and the normalized mean aggregation across all tokens. 
  }
  \label{fig:teaser}
}

\abstract{
Understanding how vision-language models (VLMs) interpret data visualizations remains an open problem, and is increasingly important as these models are used for analytical tasks where reliable reasoning is essential. We introduce a lightweight, diagnostic saliency map method tailored for text generation over images using transformer models, the current state-of-the-art models in visualization interpretation. 
Our approach aggregates the language model's attention over the visual tokens across all heads and layers, then maps this attention back onto the vision encoder's patch grid to localise it over the image, producing a direct correspondence between each generated answer token and the image regions it attended to.
This yields fast, gradient-free saliency maps that expose how VLMs allocate focus across visual elements during answer generation, enabling inspection of whether model attention aligns with semantically relevant components. We evaluate our approach using a deletion metrics which validates the causal faithfulness of our saliency maps to the model’s behavior.

} 


\begin{document}


\firstsection{Introduction}

\maketitle

Vision-language models (VLMs) are increasingly being deployed as analytical engines for data visualization-related tasks, such as querying charts, summarising trends, and extracting values from complex figures. However, the fundamental question of whether these models actually focus on the correct visual elements when answering remains less rigorously addressed. A model that consistently returns the correct number whilst attending to the wrong bar can hardly be considered a reliable tool.

Current benchmarks for Chart Question Answering (CQA) \cite{masry_chartqa_2022, methani_plotqa_2020, hutchinson_chart_2025} measure only textual correctness, leaving this blind spot unaddressed. Recent research reveals that state-of-the-art VLMs maintain surprisingly high accuracy even when chart images are corrupted or removed entirely \cite{hong2025llms}. This indicates that parametric memory and superficial textual biases often serve as shortcuts, underpinning many superficially correct answers \cite{dancette_beyond_2021}. Accuracy alone, therefore, is an insufficient proxy for visual reasoning.

In light of these limitations, we present a lightweight, gradient-free saliency method designed specifically for text generation over chart images in transformer-based VLMs. For each generated output token, our approach aggregates the language model's attention over the visual tokens across all heads and layers, and maps the resulting distribution onto the vision encoder's patch grid. This yields a per-token saliency map that reveals \emph{which pixels the model attended to} whilst producing each word of its answer without any gradient computation, architectural modification, or additional inference pass beyond standard generation

In this paper we contribute a token-level, attention-guided saliency extraction method tailored to the ViT-LLM architectures prevalent in contemporary VLMs, producing spatially coherent maps over chart images during free-form answer generation and a deletion-based faithfulness evaluation on Mini-VLAT \cite{pandey_mini-vlat_2023}, demonstrating the causal validity of our maps by showing that the removal of top-ranked pixels induces a rapid collapse in model accuracy.

\section{Related Work} 


Research into visualization literacy has historically focused on evaluating people’s ability to read and interpret data visualizations. Instruments such as the VLAT \cite{lee_vlat_2017} and CALVI \cite{ge_calvi_2023} have been created to do this. Recent work has repurposed these datasets to evaluate the state-of-the-art VLMs \cite{bendeck_empirical_2024, dong2025probing, hong2025llms}. A parallel line of work has explored large-scale benchmarking datasets for evaluating VLMs on tasks such as chart question answering \cite{masry_chartqa_2022, methani_plotqa_2020, hutchinson_chart_2025} and chart captioning \cite{tang_vistext_2023}. These benchmarks have spurred the development of specialized architectures, including ChartGemma \cite{masry_chartgemma_2025} and ChartLlama \cite{han2023chartllama}, which are fine-tuned specifically to handle the structural syntax of visualizations. Furthermore, recent work has also focused on identifying misleading visualizations \cite{lo_how_2025}; or assessing the ability to explain or justify visual encoding choices \cite{hutchinson_capturing_2025}. However, these benchmarks only evaluate whether a model outputs the correct answer, without addressing whether it arrived there by using the correct visual evidence. 
Recent studies have shown that some VLMs maintain performance even when chart images are corrupted or removed \cite{hong2025llms}. This makes accuracy alone a weak indicator of visual reasoning, and motivates the use of explainability techniques for interpreting model behavior.

Explainability has long been a central research area in computer vision and deep learning, particularly saliency-based methods that reveal which input features in an image drive a model’s decision. 
A large catalogue of saliency methods exist. Model-agnostic approaches, such as LIME \cite{ribeiro_why_2016}, approximate local decision boundaries via perturbing the image, whereas gradient-dependent approaches, such as the Grad-CAM \cite{selvaraju2017grad}, trace calculate the importance of pixels in the image through the network’s computational graph. 
Prior work has applied AG-CAM \cite{leem_attention_2024}, a version of Grad-CAM adapted for the transformer architecture, to visualization literacy explainability \cite{dong2025probing}, while others have explored aligning model-generated saliency maps with human gaze patterns \cite{salamatian_chartgaze_2025}.


Despite their popularity, existing saliency methods rely on assumptions that do not hold for modern VLMs. Techniques such as Grad-CAM \cite{selvaraju2017grad} were developed for CNNs, where spatial locality and hierarchical feature maps provide a meaningful basis for gradient-based attribution. Transformer-based architectures do not share these structural priors, instead representing images as patch tokens and propagating information globally through attention. Consequently, applying canonical saliency methods to ViTs can yield misleading explanations \cite{chefer2021transformer}. 
Moreover, most saliency methods are designed for classification tasks on natural images, whereas visualization understanding in VLMs typically involves open-ended text generation.
This means that explainability techniques for VLMs must reflect both the structure of the underlying model and the task being completed. 

In this work, we specifically address this limitation. Our framework introduces an attention-guided approach that leverages the intrinsic token-to-token attention patterns in these models, providing saliency maps that align with the mechanisms through which VLMs process visual inputs and generate open-ended answers that are grounded in the task of chart question answering.

\section{Preliminaries: Vision-Language Alignment}
In this section, we cover the necessary background for the exposition of our  proposed methodology. We focus on the prevalent paradigm used by VLMs popular  within visualization research, such as LLaVA \cite{liu_improved_2024} and  ChartGemma \cite{masry_chartgemma_2025}. These architectures couple a  pre-trained visual encoder with a language model via a trainable projection  adapter. In LLaVA, both the visual encoder and the LLM are frozen during  training and only the adapter is updated; in ChartGemma, the LLM is  additionally fine-tuned on chart-specific data.

\noindent
\textbf{Visual Encoding and Discretization}
Let the input be an image $\mathbf{I} \in \mathbb{R}^{H_{\text{img}} \times W_{\text{img}} \times 3}$, where $H_{\text{img}}$ and $W_{\text{img}}$ are assumed to be exact multiples of the patch size $P$ (achieved in practice via resizing or padding prior to encoding). The visual encoder transforms this signal into a sequence of patch embeddings. A Vision Transformer (ViT)~\cite{dosovitskiy2020image, radford2021learning} serves as the visual backbone, treating the image as a flat sequence of tokens rather than imposing the local spatial inductive bias of CNNs. The image is partitioned into fixed-size patches $\mathbf{x}_p \in  \mathbb{R}^{P \times P \times 3}$, which are flattened and linearly projected into the encoder's latent space. A CLS token $\mathbf{x}_{\text{cls}}$ is  prepended to this sequence. The encoder then produces a sequence of embeddings:
\begin{equation}
    \mathbf{Z}_v = f_{\text{enc}}(\mathbf{I}) \in \mathbb{R}^{(N_p + 1) \times D_v},
\end{equation}
where $N_p = (H_{\text{img}} \cdot W_{\text{img}}) / P^2$ is the number of patches, the additional token accounts for the CLS token, and $D_v$ is the encoder's hidden dimension. In our saliency method we consider only the $N_p$ patch tokens, excluding the CLS token, as these correspond directly to spatial regions of the image.

\noindent
\textbf{Cross-Modal Alignment via Projection}
The visual embeddings $\mathbf{Z}_v \in \mathbb{R}^{(N_p+1) \times D_v}$ and the LLM's token embeddings occupy disjoint spaces; in general $D_v \neq D_{\text{llm}}$, so direct concatenation is not possible. A trainable projection adapter $\phi$ maps the patch embeddings into the LLM's embedding 
space. In LLaVA, $\phi$ is a Multi-Layer Perceptron:
\begin{equation}
    \mathbf{H}_v = \phi(\mathbf{Z}_v) \in \mathbb{R}^{N_p \times D_{\text{llm}}}.
\end{equation}
The resulting visual tokens $\mathbf{H}_v$ act as a \emph{visual prefix}, 
conditioning subsequent text generation on the image content.

\noindent
\textbf{Autoregressive Generation and Attention Dynamics}
The input to the LLM is a unified multimodal sequence $\mathbf{X}_{\text{in}}$,  formed by concatenating the projected visual tokens $\mathbf{H}_v$ with the tokenised user query $\mathbf{X}_q \in \mathbb{R}^{N_q \times D_{\text{llm}}}$:
\begin{equation}
    \mathbf{X}_{\text{in}} = [\mathbf{H}_v;\, \mathbf{X}_q] \in 
    \mathbb{R}^{T \times D_{\text{llm}}},
\end{equation}
where $T = N_p + N_q$ is the total sequence length.

The LLM processes $\mathbf{X}_{\text{in}}$ through a stack of Transformer layers, each employing Multi-Head Self-Attention (MSA). For attention head $h$ in layer $l$, the queries, keys, and values are obtained via learned linear projections of the input: $\mathbf{Q} = \mathbf{X}_{\text{in}}\mathbf{W}^Q$, 
$\mathbf{K} = \mathbf{X}_{\text{in}}\mathbf{W}^K$, $\mathbf{V} = 
\mathbf{X}_{\text{in}}\mathbf{W}^V$. The attention matrix is then:
\begin{equation}
    \mathbf{A}^{(l,h)} = \text{softmax}\!\left(\frac{\mathbf{Q}\mathbf{K}^\top}
    {\sqrt{d_k}}\right) \in \mathbb{R}^{T \times T},
\end{equation}
where $d_k$ is the per-head key dimension. The element $A_{i,j}^{(l,h)}$ 
quantifies how much token $i$ attends to token $j$ in layer $l$, head $h$.

Answer generation is autoregressive: at each step $t$, the model produces the 
next token $y_t$ conditioned on the projected visual tokens and all previously 
generated text,
\begin{equation}
    p(y_t \mid \mathbf{H}_v,\, y_{<t}) = \text{LLM}(\mathbf{H}_v,\, y_{<t}).
\end{equation}
It is within the attention weights $\mathbf{A}^{(l,h)}$, specifically the 
attention from the token at position $t$ back to the visual prefix 
$\mathbf{H}_v$,  that we seek to locate the visual evidence underlying the 
model's decisions. The following section describes our methodology.



\section{Methodology: Attention-Guided Saliency Extraction}
\label{sec:methodology}




Our objective is to derive a spatial saliency map $S_t$ that quantifies the  influence of specific image regions (patches) on the generation of a target  text token $y_t$. We rely on a perturbation-free approach based on attention  aggregation~\cite{abnar2020quantifying}, tailored for multimodal sequences.

\noindent
\textbf{Notation.} We define the structural parameters of the model as follows:
$N_l \in \mathbb{N}$ is the total number of transformer layers;
$N_h \in \mathbb{N}$ is the number of attention heads per layer;
$N_p \in \mathbb{N}$ is the number of vision tokens (patches);
$N_q \in \mathbb{N}$ is the number of question (text prompt) tokens;
$T = N_p + N_q \in \mathbb{N}$ is the length of the input sequence
(the visual prefix followed by the question); and
$G = \sqrt{N_p}$ is the grid dimension of the vision encoder, assuming a square
$G \times G$ patch layout.
The input tokens occupy sequence positions $\{0,\dots,N_p-1\}$ for the visual
prefix and $\{N_p,\dots,T-1\}$ for the question, with generated tokens following
at positions $\geq T$.
We write $y_1,\dots,y_M$ for the $M$ generated answer tokens, and let
$t \in \{1,\dots,M\}$ index the \textit{generated} text token we wish to explain (equivalently,
the decoding step that produces $y_t$).

\noindent
\textbf{Attention Aggregation.} Transformers distribute information processing across $N_h$ heads, which often specialise in distinct linguistic features. Relying on a single head or the final layer alone yields noisy and incomplete attribution. To obtain a robust signal of visual importance, we aggregate attention weights across the full depth and width of the (language related) network. Let $\mathbf{A^{(t,l,k)}}$ denote the raw attention matrix of head $k$ in layer $l$ at decoding step $t$. We first average over heads to neutralise head-specific biases,

\begin{equation}
  \mathbf{\bar{A}}^{(t,l)} = \frac{1}{N_h}\sum_{k=1}^{N_h} \mathbf{A}^{(t,l,k)},
\end{equation}
and then over layers to obtain a single attention map for that step,
\begin{equation}
  \mathbf{A}^{(t)}_{\text{global}} = \frac{1}{N_l}\sum_{l=1}^{N_l} \mathbf{\bar{A}}^{(t,l)}.
\end{equation}

While more complex rollout strategies exist (consider~\cite{abnar2020quantifying} for a thorough exposition), we find that mean aggregation provides a stable and inexpensive signal of visual importance.

\noindent
\textbf{Vision Token Extraction.}
Rows of $\mathbf{A}^{(t)}_{\text{global}}$ are indexed by sequence position, not by
generated-token index. Let $r_t$
denote the query position whose hidden state predicts $y_t$: $r_t = T + t - 1$.
We extract this query's attention over the visual prefix, yielding the raw attention vector:
\begin{equation}
  \mathbf{w}_t = \mathbf{A}^{(t)}_{\text{global}}\big[\,r_t,\; 0\!:\!N_p\,\big] \in \mathbb{R}^{N_p}.
\end{equation}


Since {$\mathbf{A}^{(t)}_{\text{global}}$ is row-normalised over the attended keys,
$\mathbf{w}_t$ does not in general sum to 1 over the visual tokens alone. We
therefore re-normalise to obtain a distribution over patches:
\begin{equation}
    \tilde{\mathbf{w}}_t = \frac{\mathbf{w}_t}{\sum_{j=0}^{N_p-1} w_{t,j}},
,
\end{equation}
so that $\tilde{\mathbf{w}}_t$ represents the model's relative allocation of
attention across image patches, conditional on attending to the visual prefix.

\noindent
\textbf{Topological Reconstruction.}
The vector $\tilde{\mathbf{w}}_t$ is defined over a flattened patch sequence. We recover the spatial structure by mapping each patch index $i$ to its position $(r, c)$ in the $G \times G$ grid:
\begin{equation}
    r = \left\lfloor \frac{i}{G} \right\rfloor, \qquad c = i \bmod G,
\end{equation}
and reshaping to form the coarse 2D attention map:
\begin{equation}
    \mathbf{M}_t = \text{Reshape}(\tilde{\mathbf{w}}_t,\, (G, G)) 
    \in \mathbb{R}^{G \times G}.
\end{equation}
This map represents the raw attention per patch.


\noindent
\textbf{Normalisation and Spatial Upsampling.}
$\mathbf{M}_t$ is defined at patch resolution, which is substantially coarser than the input image. To produce a heatmap suitable for overlaying on the original visualization, we first apply max-normalisation to $\mathbf{M}_t$, rescaling values to $[0, 1]$ relative to the most attended patch. This is purely a display-level operation (for rendering better visualizations) that preserves the relative importance of regions across tokens while removing dependence on absolute attention magnitude:
\begin{equation}
    \hat{\mathbf{M}}_t = \frac{\mathbf{M}_t}{\max_{r,c}\, M_t[r,c]}.
\end{equation}
We then upsample $\hat{\mathbf{M}}_t$ to the original image resolution $(H_{\text{img}}, W_{\text{img}})$ via bilinear interpolation, which estimates the saliency value at each pixel $(x, y)$ as a weighted average of the four nearest patch values:
\begin{equation}
    S_t(x,y) = \sum_{i=0}^{G-1} \sum_{j=0}^{G-1} \hat{M}_t[i,j]\; 
    K\!\left(\frac{xG}{W_{\text{img}}} - j,\; \frac{yG}{H_{\text{img}}} - i\right),
\end{equation}
where $K(a, b) = \max(0, 1-|a|) \cdot \max(0, 1-|b|)$ is the bilinear interpolation kernel. The resulting map $S_t$ provides a pixel-wise approximation of the VLM's visual attention during generation of token $y_t$.

\section{Evaluation} 

\subsection{Experimental Setup and Baselines}
\label{sec:exp}
To validate our method, we first reproduce some of the results Dong and Crisan \cite{dong2025probing}, who evaluate the visualization literacy of VLMs using Mini-VLAT \cite{pandey_mini-vlat_2023} and apply AG-CAM \cite{leem_attention_2024} to probe the model behavior over input tokens. Specifically, we reproduce the test results obtained by two models (ChartGemma and LLaVA) to confirm that our setup aligns with previous results and to provide a baseline of their visualization literacy, from which we can then validate the faithfulness of our maps. Below describes the experimental setup and results. We then extend their study to use the full VLAT \cite{lee_vlat_2017}.

\noindent
\textbf{Datasets}
We evaluate our attention-guided saliency extraction method using Mini-VLAT \cite{pandey_mini-vlat_2023} and VLAT \cite{lee_vlat_2017}, but allowing the model to answer with open-ended text rather than multiple choices.
\noindent
\textbf{Models}
Two VLMs are used. LLaVA-1.5-7B is a general-purpose multimodal model using a CLIP-style ViT encoder fused with a LLaMA-based language model \cite{liu_improved_2024}. ChartGemma is a chart-focused VLM fine-tuned specifically for chart reasoning tasks, and uses a Gemma-family language model coupled with a ViT variant adapted for chart structure \cite{masry_chartgemma_2025}. The models differ in both their visual and linguistic backbones so they provide a meaningful comparison between a general multimodal system and a specialised chart-oriented one. For each question–image pair, the model receives the chart and the question in a standard prompt. The model then generates free-form text autoregressively. During generation, we compute a saliency map for the final token(s) that constitute the model’s predicted answer using the method detailed in section~\ref{sec:methodology}.

\noindent
\textbf{Evaluation Metrics}
Answer correctness is measured using relaxed accuracy, as is standard to evaluate results in chart question answering tasks \cite{masry_chartqa_2022, methani_plotqa_2020}. Rather than enforcing strict string matching, the VLM-generated answer is considered correct if the correct answer appears anywhere within the generated token sequence, and additionally for quantitative numerical answers, answers are considered correct if they are within 5\% of the ground truth answer. We report the accuracy over all samples as a percentage.

\noindent
\textbf{Results} On MiniVLAT, ChartGemma achieves 58.3\% accuracy on this setup, which matches exactly the result reported by \cite{dong2025probing}. LLaVA-1.5 achieves 25\%, which is slightly more than the 20\% reported by \cite{dong2025probing}, but within an acceptable range given that VLMs are not deterministic. These results confirm the validity of our generation setup. On VLAT, ChartGemma achieves 34.0\% and LLaVA-1.5 22.6\%. These values confirm that ChartGemma’s finetuning increases performance on visualization literacy tests.

\subsection{Deletion-Based Faithfulness Evaluation}
To evaluate the faithfulness of our saliency maps, we employ a deletion test, a causal metric for evaluating saliency explanations \cite{petsiuk_rise_2018}. This methodology systematically removes image regions identified as most salient by a given attribution method and measures the resulting degradation in model performance. This provides a quantitative assessment of whether the highlighted regions genuinely contribute to the model's performance, and thus whether the saliency maps themselves are faithful representations of the model's focus. We run these tests on both our maps and AG-CAM.

\noindent
\textbf{Dataset and Models} We use the same setup as \ref{sec:exp}, using VLAT. However, given that deletion measures \textit{causal} faithfulness, we only use samples which the model answered correctly in the vanilla setting. Additionally, we exclude questions which have a True/False answer, as the probability of the model ``guessing'' the right answer is high, which interferes with the meaningful signal from the deletion testing.  Given this, we only perform deletion evaluations with ChartGemma, as LLaVA did not answer enough questions correctly to provide a representative sample. Thus, we run the following evaluations using $13$ samples from VLAT and ChartGemma. Sample sizes of this order are consistent with prior deletion-based evaluations.\cite{boggust_saliency_2023, petsiuk_rise_2018}.

\begin{figure}[h]
 \centering
 \includegraphics[width=\linewidth]{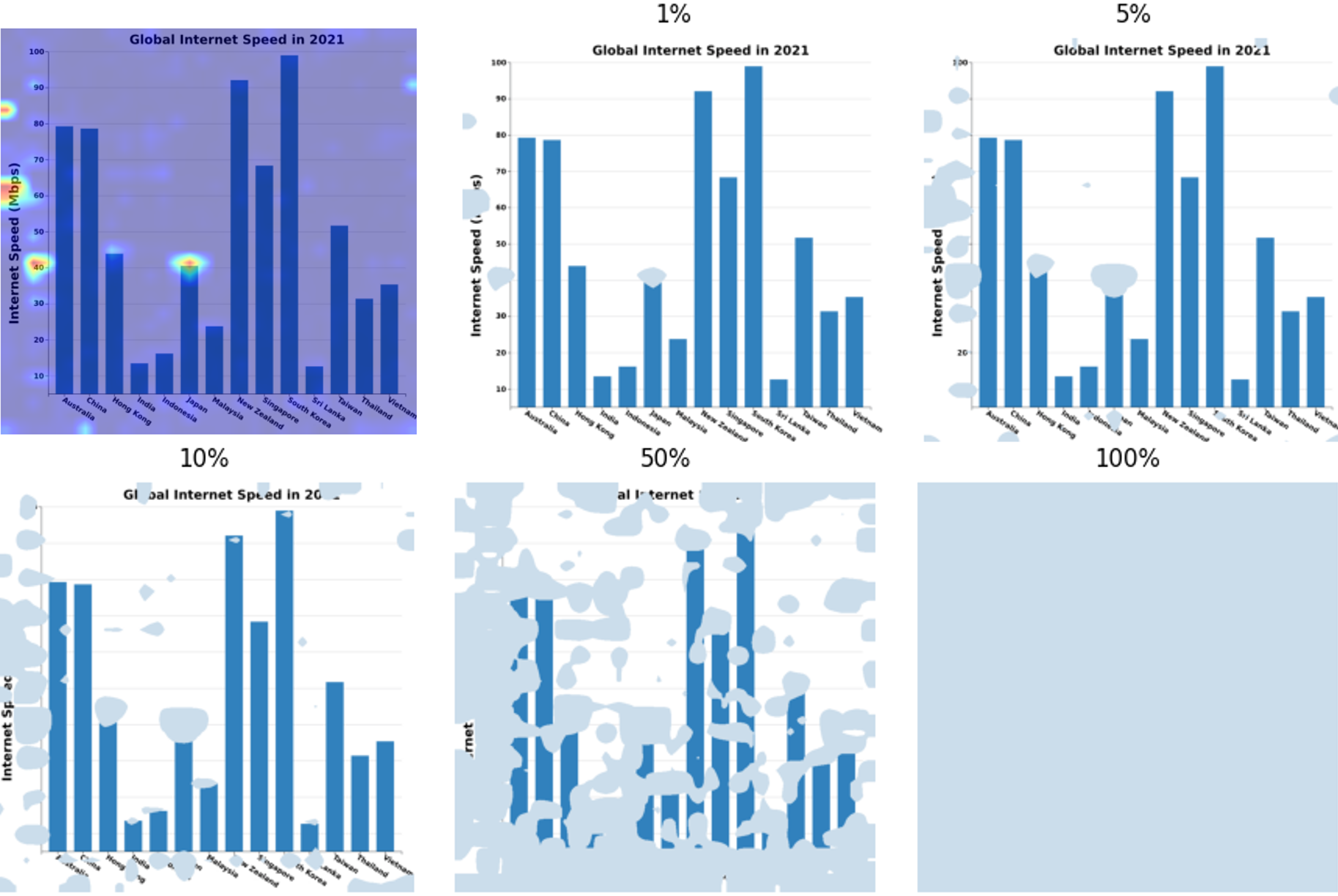}
 \caption{Progressive pixel deletion applied to a visualization from VLAT using our saliency map method. The top-left panel shows the aggregate saliency map;  subsequent panels show the chart after removing the top $k\%$ of pixels ranked by saliency.}
 \label{fig:deletion}
\end{figure}

\noindent
\textbf{Deletion Methodology} For each question–chart pair in the evaluation set, we first obtain an aggregate saliency map by running the model on the unmodified question and image. 

For our approach, we do this by first extract the attention weights over image patch tokens at each decoding step, as described in \ref{sec:methodology}. We then L1-normalize each token's attention distribution to ensure that tokens with diffuse attention patterns contribute equally to the aggregate map rather than being dominated by tokens with sharply peaked distributions. Finally, we average across all generated tokens to produce a single spatial heatmap representing the entire generated answer. The rightmost heatmaps in figure \ref{fig:teaser} show this.

For AG-CAM, we perform a forward pass followed by backpropagation from the maximum logit, compute the element-wise product of the ReLU-activated gradients and the attention weights at each layer, sum across attention heads, and accumulate across all transformer layers to obtain the final attribution map. It is worth noting that in contrast to our approach, AG-CAM operates over \textit{input} tokens rather than \textit{output} tokens generated by the model, and so the saliency maps do not represent focus during answering, but parts of the image that are salient to the question.

Given the aggregate saliency map, we rank all pixels by their attributed importance and progressively replace the top $k\%$ most salient pixels with the channel-wise mean color of the original image. We choose mean-color infill to avoid introducing out-of-distribution artifacts that solid black or gray patches might cause \cite{petsiuk_rise_2018}. We do this for all values of $k \in \{0,1,2,...,100\}$. Figure \ref{fig:deletion} shows an example our saliency map and corresponding degraded images at different values of $k$. At each deletion level, we prompt the model with the original question and degraded image, and record whether the model's answer remains correct. We report accuracy as a percentage as a function of the percentage of pixels deleted, producing an accuracy-versus-deletion curve. As a control, we also repeat the procedure with randomly selected pixels.

\noindent \textbf{Evaluation Metrics} We evaluate each accuracy-versus-deletion curve using the area under the curve (AUC) \cite{petsiuk_rise_2018}. A lower AUC for saliency-based deletion relative to the random baseline indicates that the saliency method successfully identifies pixels that are causally relevant to the model's output: removing them degrades performance more rapidly than removing an equivalent number of arbitrarily chosen pixels. 

\begin{figure}[h]
 \centering 
 \includegraphics[width=\columnwidth]{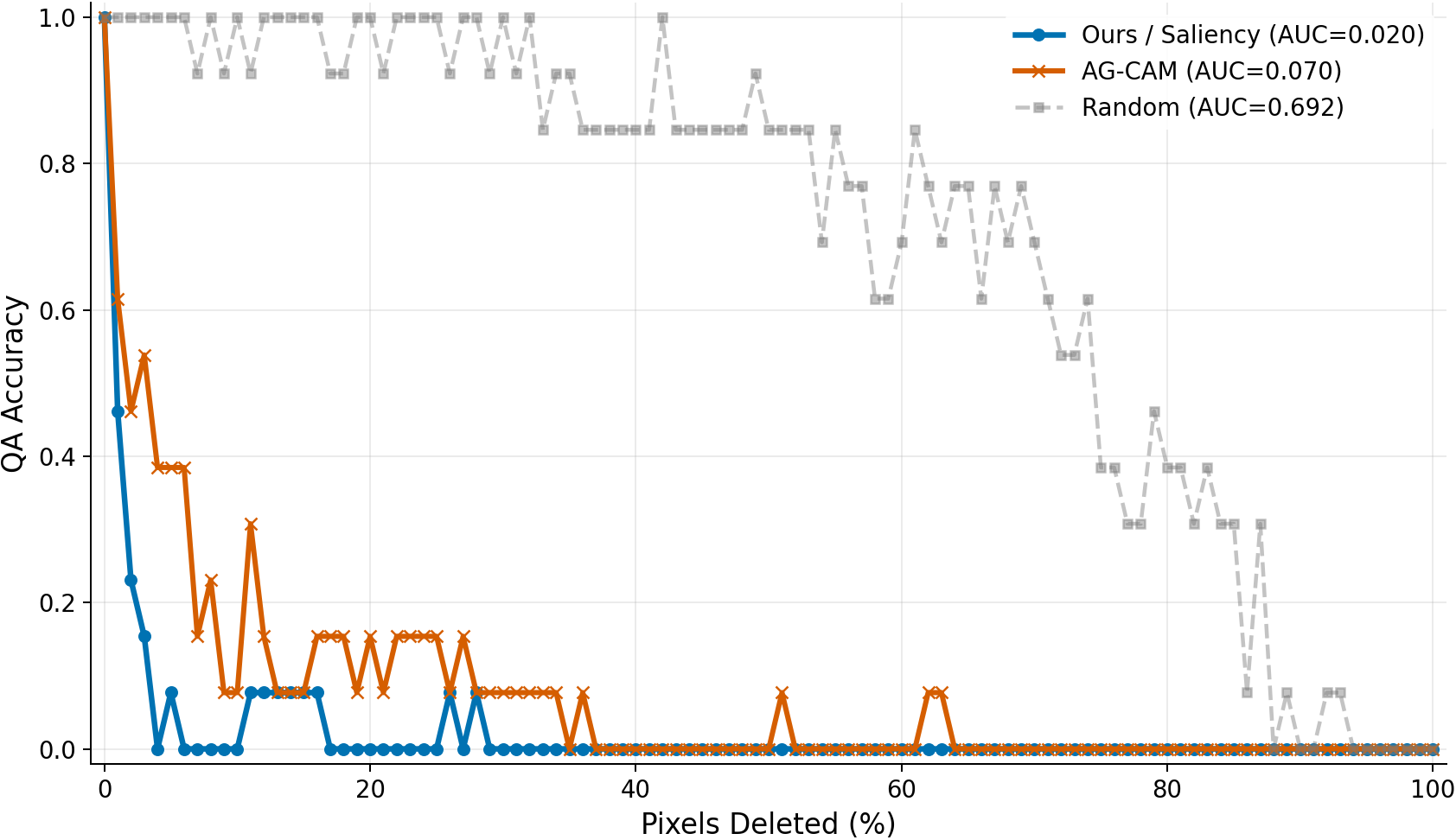}
 \caption{Deletion test results for ChartGemma on 13 VLAT samples. Accuracy is plotted against the percentage of pixels deleted for our saliency method, AG-CAM and random pixel deletion.}
 \label{fig:vis_papers}
\end{figure}

\noindent
\textbf{Results}
Figure \ref{fig:vis_papers} presents the results of the progressive deletion test for ChartGemma. Both our method and AG-CAM cause a rapid collapse in accuracy compared to random deletion, confirming that both methods identify genuinely informative regions rather than arbitrary pixels. Our method achieves an AUC of 0.020, outperforming AG-CAM at an AUC of 0.070, indicating that our approach more faithfully reflects the focus of the model. The steepness of both saliency curves also reveals that chart question answering is highly spatially concentrated: the model's ability to answer correctly depends on a very small fraction of the image, typically the specific axis label(s) or mark(s) that encode the queried value. Once these critical pixels are removed, performance collapses almost entirely, even though the majority of the visualization remains intact.

\section{Conclusion}

We have presented a lightweight, gradient-free saliency extraction method for interpreting how vision-language models process data visualizations during text generation. By aggregating attention weights across the language model's layers and mapping them back through the vision encoder's patch topology, our approach produces per-token and aggregate saliency maps that reveal which visual elements the model relies on when generating each part of its answer. 

Our deletion-based evaluation confirms that these maps are causally faithful, achieving this with a lower AUC than AG-CAM, while requiring no gradient computation, no architectural modification, and no additional inference passes beyond standard generation, and importantly being caculated over \textit{output} tokens rather than \textit{input}. This makes it practical for deployment as a real-time diagnostic tool alongside model predictions.

More broadly, this work contributes to an emerging need within the visualization community: as VLMs are increasingly used for analytical tasks, we require interpretability tools that are tailored to both the architecture of these models and the structured nature of data visualizations. Saliency maps that faithfully reveal whether a model is attending to the correct regions of a visualization provide a necessary complement to accuracy-based evaluation, helping identify not just when a model fails, but why.

\section*{Supplementary material}
The dataset is code available at \url{https://github.com/maevehutch/token-patch-saliency}.

\section*{Acknowledgments} This work
was supported, in part, by the Alan Turing Institute under Fundamental Research (Project No. PP00029).

\bibliographystyle{abbrv-doi}

\bibliography{egbibsample, eurovis-saliency, template}
\end{document}